\documentclass[letterpaper, 10 pt, conference]{ieeeconf}  

\IEEEoverridecommandlockouts                              

\usepackage{graphics} 
\usepackage{float}
\usepackage{siunitx}
\usepackage{multirow}
\usepackage{amsmath}
\usepackage{bbm}
\usepackage{amssymb}
\usepackage{mathtools}
\usepackage[utf8]{inputenc} 
\usepackage[T1]{fontenc}    
\usepackage{hyperref}       
\usepackage{url}            
\usepackage{booktabs}       
\usepackage{amsfonts}       
\usepackage{nicefrac}       
\usepackage{microtype}      

\usepackage[table]{xcolor}

\title{\LARGE \bf
ImageManip: 
Image-based Robotic Manipulation with
Affordance-guided Next View Selection
}

\author{Xiaoqi Li, Yanzi Wang, Yan Shen, Ponomarenko Iaroslav, Haoran Lu,\\
Qianxu Wang, Boshi An, Jiaming Liu, Hao Dong
\thanks{All authors are with School of CS, Peking University and National Key Laboratory for Multimedia Information Processing.
Xiaoqi Li is also with Beijing Academy of Artificial Intelligence (BAAI).
}
\thanks{
Corresponding to hao.dong@pku.edu.cn}
}

\begin{document}

\maketitle
\thispagestyle{empty}
\pagestyle{empty}

\begin{abstract}
In the realm of future home-assistant robots, 3D articulated object manipulation is essential for enabling robots to interact with their environment.  
Many existing studies make use of 3D point clouds as the primary input for manipulation policies. 
However, this approach encounters challenges due to data sparsity and the significant cost associated with acquiring point cloud data, which can limit its practicality.
In contrast, RGB images offer high-resolution observations using cost-effective devices but lack spatial 3D geometric information.
To overcome these limitations, we present a novel image-based robotic manipulation framework. This framework is designed to capture multiple perspectives of the target object and infer depth information to complement its geometry.
Initially, the system employs an eye-on-hand RGB camera to capture an overall view of the target object. 
It predicts the initial depth map and a coarse affordance map.
The affordance map indicates actionable areas on the object and serves as a constraint for selecting subsequent viewpoints. 
Based on the global visual prior, we adaptively identify the optimal next viewpoint for a detailed observation of the potential manipulation success area. We leverage geometric consistency to fuse the views, resulting in a refined depth map and a more precise affordance map for robot manipulation decisions.
By comparing with prior works that adopt point clouds or RGB images as inputs, we demonstrate the effectiveness and practicality of our method. 
In the project webpage (\url{https://sites.google.com/view/imagemanip}), real-world experiments further highlight the potential of our method for practical deployment.

\end{abstract}

\section{INTRODUCTION}
The research community has recently shown a growing interest in embodied AI and its practical applications in the field of robotic manipulation. 
Recent studies~\cite{lv2022sagci,geng2022end,geng2023rlafford,zhao2022dualafford,wang2022adaafford,wu2021vat} have highlighted the potential of using point-cloud cameras to obtain detailed 3D geometric information for manipulation tasks.
However, collecting point-cloud data involves capturing 3D depth information, which frequently leads to sparse data representation. This sparsity poses severe noises, especially on specular or transparent material surfaces with high-intensity lighting.

\begin{figure}[t]\label{Teaser}
\begin{center}
    \includegraphics[width=8cm,height=4.3cm]{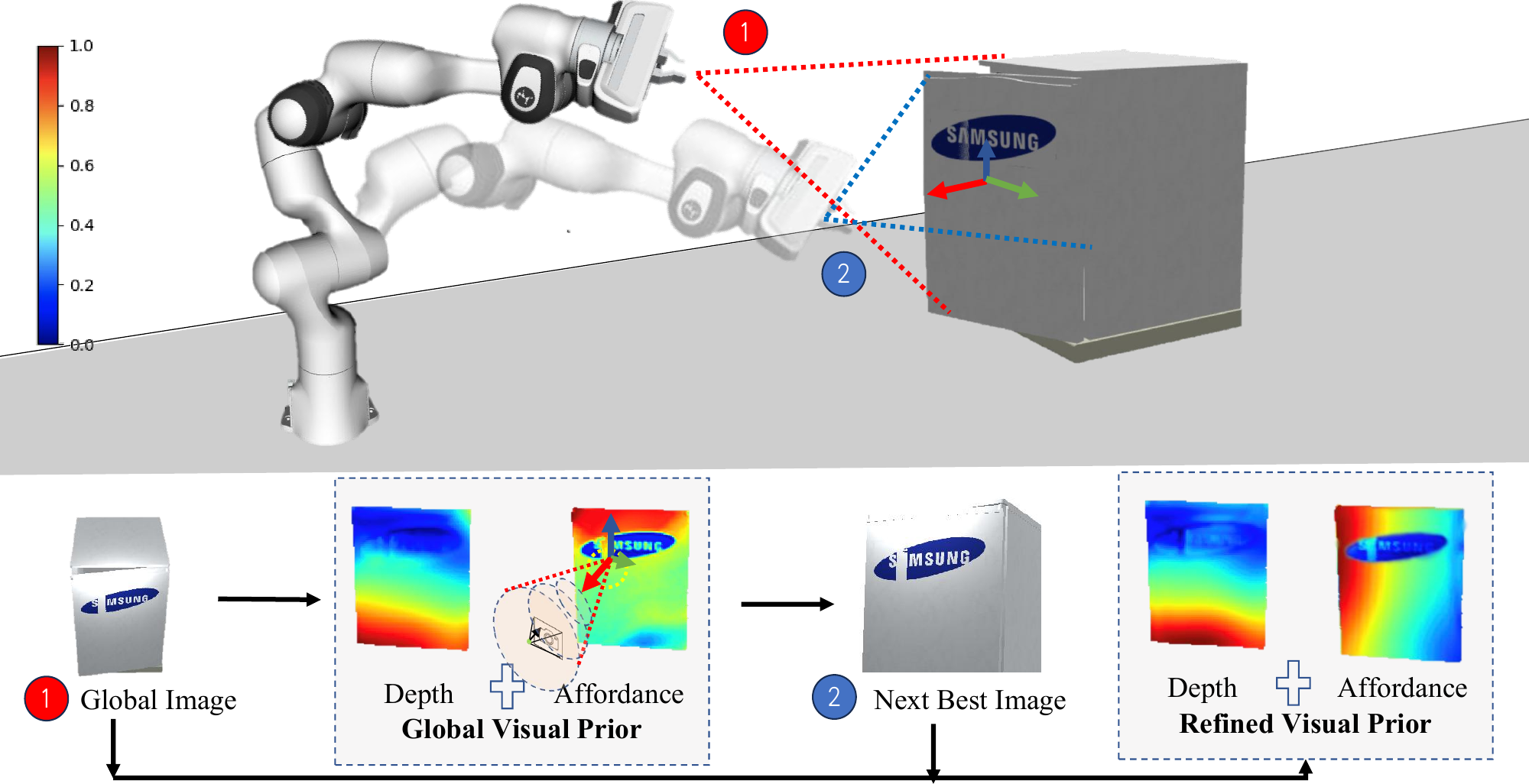}
\end{center}
\caption{The overall framework. 
In the depth map, the values represent normalized depth distances, while in the affordance map, the value of each pixel represents the probability of successful manipulation at that point.}
\vspace{-0.2cm}
\label{fig:teaser}
\end{figure}
	


To enhance the level of captured detail, people consider increasing the resolution of point-cloud data, or alternatively, adopting industrial-grade point cloud cameras, both of which can increase hardware costs and limit their practicality.
In contrast, RGB images present a readily available and cost-effective visual modality that captures detailed observations with relatively higher resolution.
They are widely employed in various computer vision tasks ~\cite{haralick1973textural,hao2020brief,dai2022graspnerf} due to their accessibility and compatibility with existing imaging systems. 
While RGB images lack inherent 3D geometry, recent advancements in the field of self-driving cars have demonstrated promising techniques to complement the geometric information in RGB images by leveraging depth estimation~\cite{li2022bevdepth,li2022bevstereo,ma2022vision}.
To harness the benefits of the RGB modality, which is characterized by its widespread availability and high-resolution capabilities, we introduce an RGB-only robot manipulation framework. 
Our approach utilizes an eye-on-hand RGB camera to capture diverse perspectives of the target object without the need for moving back and forth.
To compensate for the absence of geometric information, we incorporate depth estimation for better manipulation decisions.

Specifically, as shown in Fig. \ref{fig:teaser}, our process begins by capturing a global view image, providing an overall view of the target object.
This image is used to generate a global visual prior, which comprises an initial depth map and a coarse affordance map.
The depth map approximates the object's 3D structure, while the per-pixel affordance map identifies where the robot's end-effector should engage with the object for successful manipulation.
Recognizing the limitations of a single view, we proceed to capture additional views and fuse their features to construct a comprehensive visual representation.
We introduce a next view selection module, which dynamically determines the optimal camera position for capturing the subsequent view based on the global visual prior. 
As we obtain the next view and merge the features from different views, we generate the refined depth map, improving our estimation of the object's 3D structure. 
Simultaneously, we obtain the refined per-pixel affordance map, providing a more precise indication of the regions where the robot's end-effector should interact. 
This enhanced affordance map, in turn, serves as a supervision for the next view selection module whose goal is to select the optimal view that can provide the most extra valuable information for manipulation decisions.

We adopt the SAPIEN simulator~\cite{Xiang_2020_SAPIEN} to train and evaluate our approach for pushing and pulling on 972 shapes across 15 commonly encountered indoor object categories. 
Following the evaluation method of prior studies~\cite{Mo_2019_CVPR,eisner2022flowbot3d,xu2022universal}, our experimental results demonstrate that our method effectively learns to predict feasible actions for novel objects, even for object categories that have not been encountered during training phase.
\section{Related work}

One widely used approach in robotic control and planning is state-based reinforcement learning (RL)~\cite{joshi2020robotic,andrychowicz2020learning,yarats2021mastering,geng2022end,geng2023partmanip,geng2023gapartnet}. 
Some works have identified the possibility of using the pure state as the policy input~\cite{andrychowicz2020learning}. However, when it comes to more complex settings, vision-based observation becomes necessary to acquire fine-grained 3D geometry information.

As for vision-based robotic manipulation, numerous studies have explored the tasks, including object grasping~\cite{ichnowski2021dex,dai2022graspnerf}, articulated object manipulation~\cite{Mo_2019_CVPR,geng2022end}, and object reorientation~\cite{andrychowicz2020learning}. Various visual modalities have been explored for perceiving the environment in robotic manipulation. Some studies, such as SAGCI-System\cite{lv2022sagci}, RLAfford\cite{geng2022end}, and Flowbot3D\cite{eisner2022flowbot3d}, have employed solely point clouds as observations. On the other hand, approaches from Xu~\cite{xu2022universal} and Wu~\cite{wu2020grasp} have integrated both RGB images and point clouds for tasks like articulated object manipulation and object grasping.
However, the use of point clouds poses challenges when dealing with specular and transparent objects as they interfere with the imaging process of current depth cameras~\cite{sajjan2020clear}. 
Thus, some works have discussed the advantages of using RGB-only image inputs to achieve robust robotic manipulation~\cite{ichnowski2021dex, dai2022graspnerf} for its high pixel-wise resolution and ability to be readily accessed. In this work, we explore only RGB-based manipulation but acquire depth estimation for perceiving objects' 3D structures.

In addition, several studies have demonstrated the possibility of improving vision-based robotic manipulation via affordance prediction~\cite{Mo_2019_CVPR,wu2021vat,geng2022end}. In the work Where2Act~\cite{Mo_2019_CVPR}, affordance is defined as a per-point actionable score that measures the possibility of success when the robot interacts with each point on the object, while RLAfford~\cite{geng2022end} defines affordance as the contact frequency during the RL exploration. In this work, we follow the definition from Where2Act, but extend our consideration of affordance as an explicit constraint to narrow down the possible range of movement for the robot's end-effector during both the training and inference stages.

\section{Methodology}

\subsection{Overall Framework}
The whole procedure of the overall framework is presented in Fig. \ref{fig:teaser}.
Initially, we capture a global image $I_{1}\in \mathbb{R}^{H\times W \times 3}$, and obtains global visual prior, \emph{i.e.}, depth map $D_{1}\in \mathbb{R}^{H\times W \times 1}$ and affordance map $A_{1}\in \mathbb{R}^{H\times W \times 1}$. 
The depth map complements the lack of object's 3D geometry, while the per-pixel affordance map indicates each pixel's ``actionability'' in accomplishing the manipulation task.

\begin{figure}[H]
\includegraphics[width=8.3cm, height=4cm]{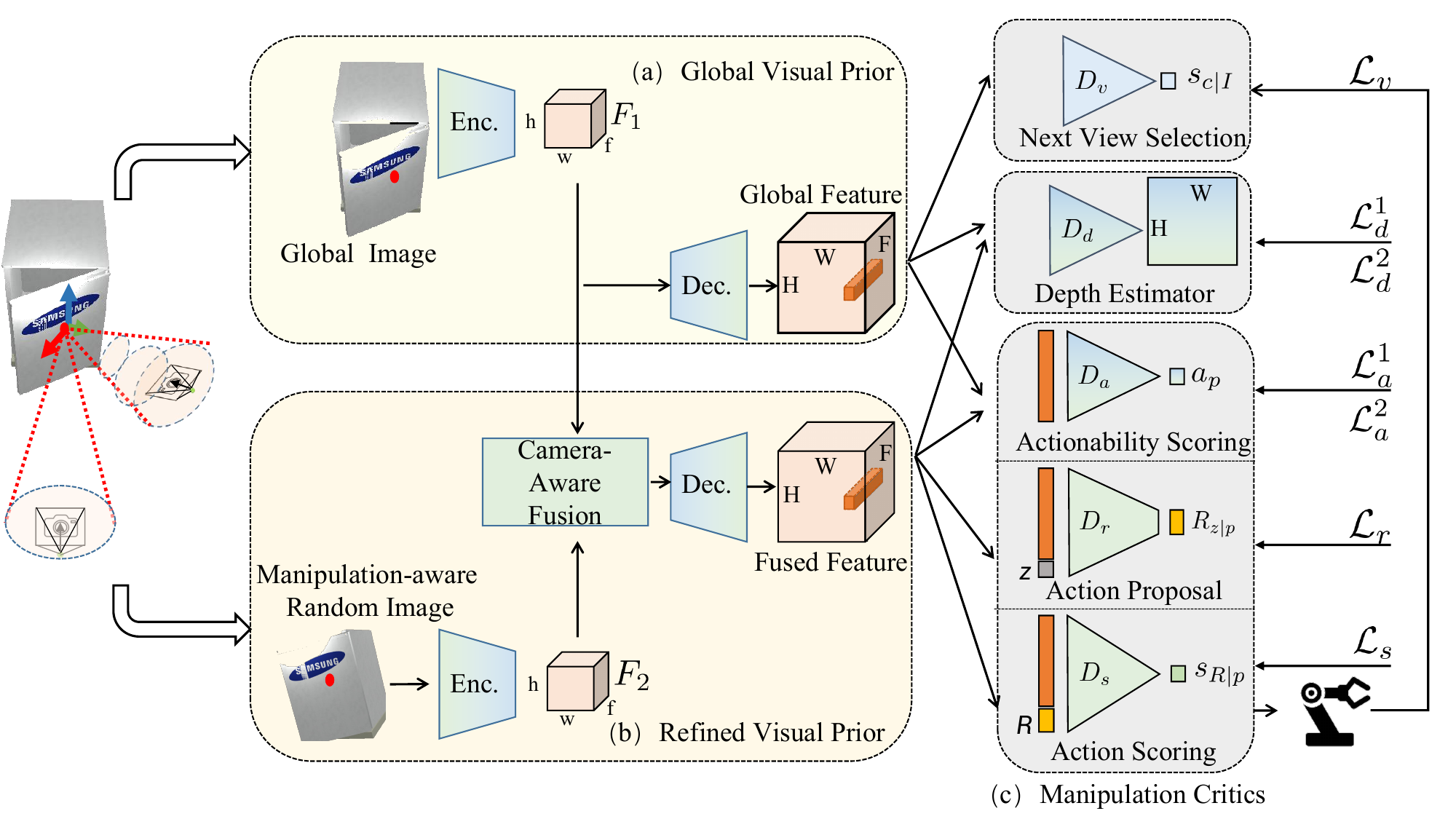}
\centering
\caption{
Training Strategy for the Entire Network. Modules in blue only take feature from global image while modules in green only take fused feature. Modules in the mix of blue and green are shared for both views. The Orange cube denotes the feature of the contact point.
}
\label{fig:overview}
\end{figure}

Based on the global visual prior, the \textit{next view selection} selects the optimal camera position to capture the best next image $I_{2}\in \mathbb{R}^{H\times W \times 3}$.
Subsequently, in \textit{camera\_aware fusion} module, the feature of $I_{1}$ and $I_{2}$ are token-wise fused to generate the refined visual prior, \emph{i.e.}, refined depth map $D_{2}\in \mathbb{R}^{H\times W \times 1}$ and affordance map $A_{2}\in \mathbb{R}^{H\times W \times 1}$. 
We then select contact point $p$ with the highest ``actionability" score on the refined affordance map. 
Apart from this, we further generate end-effector's orientation with action proposal and action scoring module.
Practically, during training, the data is pre-collected in a random manner. Following the training procedure in Fig. \ref{fig:overview}, we train all modules simultaneously. 

\subsection{Obtaining Global Visual Prior}
Given the global image, as shown in Fig. \ref{fig:overview}(a), we aim to obtain global visual prior, \emph{i.e.}, \textbf{initial depth map} and \textbf{initial affordance map}.
The goal of global visual prior is to serve as the condition for the following next view selection and provide global observation for the manipulation policy.

\textbf{Initial Depth Map.} 
Given the global image $I_1$ with $H$ height and $W$ width, 
we extract high-dimensional per-pixel features $F_1$ with an encoder. 
The encoder employed is based on the architecture of U-Net\cite{ronneberger2015u}, featuring a ResNet-18~\cite{he2016identity} encoder that has been pre-trained on ImageNet~\cite{deng2009imagenet}.
We then obtain the global feature with a decoder that is trained from scratch. 
Additionally, dense skip links are equipped between the encoder and decoder. 
We then adopt the depth estimator $D_{d}$ to convert global feature to initial depth estimation map $D_{1}\in\mathbb{R}^{H\times W\times1}$, which contains an MLP\cite{murtagh1991multilayer} to aggregate channel-wise representation.
It is then supervised by depth ground truth under L1\cite{carvalho2018regression} loss $\mathcal{L}_{d}^{1}$. 
The goal of the initial depth map is to help place the camera in 3D space to capture the following next view.
 
\textbf{Initial Affordance.} 
The initial affordance map aims to screen out the region of manipulation interest and enable the subsequent views to focus on this region, thus reducing the exploration space for the following manipulation steps.

To obtain the affordance map, given the global feature, we adopt the actionability scoring module $D_{a}$ to measure the actionability of each pixel. 
It contains a multi-layer perception (MLP) and a sigmoid function and converts global feature of the given pixel to probability $a_{p}\in [0,1]$, indicating the likelihood of the pixel being operable.
We supervise $D_{a}$ with the ground truth of 1 or 0, representing whether the manipulation on this point is a successful interaction. 
The pixel and the corresponding manipulation success result are collected beforehand.
The loss $\mathcal{L}_{a}^{1}$ is a standard binary cross-entropy loss\cite{ruby2020binary}.
By projecting this 2D affordance with the initial depth map, we can estimate the region of manipulation interest in 3D space. Subsequently, after identifying the next view $I_{2}$ based on the global visual prior, we integrate its feature with the global view feature to formulate a comprehensive visual representation for enhancing depth estimation and affordance prediction (Sec~\ref{sec:fusion}).
As for the process of determining the next view, we adaptively choose the next view to maximize the chances of predicting successful manipulation (Sec~\ref{sec:view}).

\subsection{Geometric-consistency Image Fusion}
\label{sec:fusion}
To enhance the model's ability to capture a broader visual context, we introduce geometric-consistency fusion, which combines the feature of global view and the subsequent view. However, this fusion process faces challenges due to differences in viewing angles between the two views, leading to issues like foreground obscure. Therefore, it's vital to ensure that both views share common information and allow us to fuse.
To address this, we first identify an initial contact point, denoted as $p_{1}$, which has the highest score on the initial affordance map $A1$. We use this point as the center of the camera and place the camera into the 3D space with the help of initial depth map to capture the next view $I_{2}$. This approach guarantees that there is at least some overlap between the regions near point $p_{1}$ that can be fused.
This region is also of high investigation interest since it is of high manipulation success probability as predicted by the initial affordance map. Though inaccuracy may exist in depth prediction that may impact the location to place the camera, it will not influence its orientation and we still can ensure it is with $p_{1}$ as camera center. By doing so, we ensure the movement of the robot end-effector is directed towards a region that is very likely to be manipulated, thus avoiding introducing additional unnecessary movement.

Next, to find the pixel-wise correspondence between the two views, we map $I_{1}$ into 3D space using the initial depth map, $D_{1}$, and project it onto the camera coordinate of $I_{2}$.
However, this process may contain misalignment due to inaccuracies in depth prediction ($D_{1}$), causing error accumulation in pixel-wise fusion process.
To avoid such, we propose to perform fusion at a token-wise level after the encoding stage. This token-wise fusion occurs at a lower resolution in which one token aggregates the feature of n$\times$n pixels, and n is the dimension of the high-dimensional feature. 
Even if there are issues with pixel correspondence, as long as the corresponding pixels fall within the token composed of n$\times$n pixels around the correct corresponding point, no additional errors will be introduced.
It is more tolerant of inaccuracies in depth estimation.

Specifically, we determine token-level correspondence based on pixel-level correspondence, by collecting all pixel correspondences within a token and then selecting the most relevant token.
In Fig. \ref{fig:overview}(b), we first extract the high-dimension feature $F_{2}$ of the next view with the same encoder, and then fuse the two high-dimension features based on the token-wise correspondence. The fused token has the same semantic representation.
To transfer the next view to the global view, in $F_{1}$ and $F_{2}$, if a token from F1 has a corresponding token in $F_{2}$, we merge the two. If a token from F1 does not have a corresponding token in $F_{2}$, we retain the token from $F_{1}$.
After obtaining the fused feature, the same actionability scoring module $D_{a}$ and depth estimator $D_{d}$ are adopted to predict refined depth $D_{2}$ and affordance map $A_{2}$ under loss $\mathcal{L}_{a}^{2}$ and  $\mathcal{L}_{d}^{2}$. We then project the affordance map into 3D space with the depth map and select the point with the highest affordance score as the contact point to interact with the object.
We visualize pixel-wise and token-wise correspondences in the project webpage. We also provide experiments in Sec.~\ref{sec:abla} to show the effectiveness of  token-wise fusion in tolerating inaccurate depth prediction.

\subsection{Next View Selection}
\label{sec:view}
This section is focused on selecting the viewpoint to increase the chances of predicting a successful manipulation. 
By analyzing the movement of the eye-on-hand camera, we notice that most views at different time stamps are similar and do not offer important additional information, resulting in unnecessary redundancy if we leverage all of them. 
Therefore, this selection mechanism aims to not only reduce the number of incorporated views by only selecting useful views but also ensure efficient manipulation moving. 
We first illustrate the process of selecting the optimal next view and then demonstrate how to supervise the selection module.

\textbf{Best Next View Selection Module.} 
During inference, the robot cannot explore randomly to assess various viewpoints. Hence, in our training process, we train a next view proposal network to adaptively choose the next image, denoted as $I_{2}$, based on the prior global view.

Specifically, during training, after capturing $I_{1}$, we employ a random data collection method inspired by Where2Act\cite{Mo_2019_CVPR}. 
We randomize the contact point $p$ for object interaction and divide the surrounding 3D space into several candidate areas. The camera is then randomly placed in one of these candidate areas with $p$ as the camera's center to capture the next view, $I_{2}$. 
The best view selection module, denoted as $D_{v}$, learns to evaluate the camera position to capture the next view, which is conditioned on the  global feature.
With the trained view selection module, during inference, and based on the predicted initial affordance map, we can project it into 3D space with the assistance of the initial depth map and roughly identify the region of manipulation interests. We apply the same 3D space division strategy as used in training to obtain candidate viewpoints. 
The next view selection module is then employed to select from all candidates and determine the optimal camera pose for capturing $I_{2}$. This approach empowers the robot to make informed decisions about the next viewpoint.
Also, by placing the camera towards the region that worth investigating, we ensure efficient movement.

\textbf{Supervision for the Next View Selection.}
Based on the global view features and camera pose information, the view selection module outputs the probability that the next view captured at the given camera pose is a good view.
During training, we introduce supervision based on the confidence score improvement when predicting affordance ($A_{2}$) compared to the initial affordance estimation ($A_{1}$). This approach helps us choose the next view that can provide a more accurate affordance prediction, thereby enhancing the overall visual feature representation and manipulation prediction.
Specifically, the actionability scoring module is a binary classification module that assigns a confidence score to its prediction on whether interacting with a given pixel will result in a successful manipulation. 
When collecting ground-truth, we assign a value of 1 to a pixel if $c_{2}$ is greater than $c_{1}$; otherwise, we assign -1. $c_{1}$ and $c_{2}$ represent the confidence scores for affordance predictions for a given pixel.
We calculate the average score for all pixels in the images.
The next view is considered valuable if the average score is greater than 0, indicating that the confidence score for affordance prediction has increased for over 50\% of the pixels. 
We use a binary cross-entropy  loss\cite{ruby2020binary} ($\mathcal{L}_{v}$) for supervision.


\subsection{Manipulation Planning}
\textbf{Manipulation policy.}
Inspired by Where2Act\cite{Mo_2019_CVPR}, to identify the manipulation orientation, in Fig. \ref{fig:overview}(c), we adopt an \textbf{action proposal module} $D_{r}$ to propose action proposals given the fused feature of contact point $p$, which is selected from the refined affordance map $A_{2}$. It contains an MLP to predict end-effector 3-DoF orientation $R_{z|p}\in SO(3)$, z is drawn from a Gaussian distribution. We train $D_{r}$ to be able to propose one candidate that matches the ground-truth interaction orientation, which is penalized by the distance function between two 6D-rotation representations $\mathcal{L}_{r}$.
Meanwhile, to determine one action proposal from all candidates, we adopt the \textbf{action scoring module} $D_{s}$ to evaluate the manipulation success rate $s_{R|p}\in [0,1]$ of implementing a predicted action proposal $R$, given the fused feature of the contact point $p$. We use $D_{a}$ to supervise it because they are positively correlated. Specifically, we calculate the sum of 100 likelihoods of $D_{s}$ on 100 candidate proposals given a pixel feature $f_{p}^{2}$. It is then supervised by the prediction of $D_{a}$ under MSE loss $\mathcal{L}_{s}$. After adjusting the relative loss scales to the same level, we obtain the total objective function: $\mathcal{L}$ = ($\mathcal{L}_{a}^{1}$ + $\mathcal{L}_{a}^{2}$) + $\mathcal{L}_{r}$ + $100*\mathcal{L}_{s}$ + ($\mathcal{L}_{d}^{1}$ + $\mathcal{L}_{d}^{2}$) + $\mathcal{L}_{v}$. All modules are trained simultaneously under this total objective $\mathcal{L}$.

\textbf{Force-feedback Closed Loop Adjustment.} 
So far, we have obtained the contact point from refined affordance map and orientation of the gripper (p,R), to realize the first interaction with the object.
In our eye-on-hand setting, we face limitations with vision-based closed-loop physical manipulation. 
When attempting to manipulate objects after the first step interaction, the gripper would get too close, resulting in insufficient visual information for generating subsequent predicted (p,R) pairs. 
To overcome this challenge, we adopt a different approach that relies on acquiring information from the force applied to the robotic arm without moving the camera back and force \cite{whitney1977force}.
The control policy leverages the target pose, current pose, and previous pose to compute an intermediate pose, which is then fed into an impedance controller\cite{hogan1985impedance}. This controller generates a sequence of pairs (p,R) and effectively steers the robotic arm towards the intermediate pose, accomplishing the long-term manipulation step by step. 
This method allows us to develop a reliable manipulation policy that is resistant to disturbances and capable of handling both revolute and prismatic articulated objects.

\section{Experiment}

\subsection{Experiment Setting}

\textbf{Data collection}
We adopt SAPIEN~\cite{Xiang_2020_SAPIEN} and the PartNet-Mobility dataset to set up an interactive environment for our task, with VulkanRenderer of high-efficiency rasterization-based renderer.
In order to tackle the significant disparity between visual and collision shapes, we leverage the V-HACD \cite{mamou2016volumetric}(Voxelized Hierarchical Approximate Convex Decomposition) algorithm.
This method entails voxelizing the 3D model, subsequently engaging hierarchical approximation to iteratively diminish the voxel count and amalgamate them into larger convex voxels. Subsequently, convex decomposition is applied to transform these merged convex voxels into simpler convex shapes.
We use a Franka Panda Flying gripper with two fingers as the robot actuator.
We consider two types of action primitives pushing and pulling with pre-programmed interaction trajectories. 
Following Where2Act\cite{Mo_2019_CVPR}, we sample the training data in an offline fashion with approximately 1000 positive and 1000 negative trajectories per category.
We randomly sample an interaction orientation $R\in SO(3)$ from the hemisphere above the tangent plane around $p$, oriented consistently to the positive normal direction, and collect the outcome of the interaction given ($p$, $R$) in the simulator.
As for the next view collection, the 3D space to place the camera for capturing next view is limited to a distance between 2.5 and 4.5 units from $p$, an azimuth angle ranging from $-\ang{20}$ to $\ang{+20}$, and an altitude angle ranging from $-\ang{20}$ to $\ang{+20}$ around the normal of that point.
We split the space into nine candidates and randomly set camera in one candidate to capture the next view.

\textbf{Evaluation metric.}
We adopt the manipulation success rate to reflect the outcome of the manipulation which is the ratio of the number of successfully manipulated samples divided by the total number of all test samples. As for the definition of success sample, we adopt the binary success definition which is measured by thresholding the move distance of the object part at $\delta$: success = $1(\delta_{dis}>\delta)$.
Following Where2Act\cite{Mo_2019_CVPR} and Flowbot3D\cite{eisner2022flowbot3d}, we set $\delta = 0.01$ or $\delta = 0.1$ for short-term `atomic' or long-term interaction respectively, meaning that the gap between the start and end part 1-DoF pose is greater than 0.01 or 0.1 unit-length. 
As for the metric to measure the performance of the depth estimation module, we adopt the commonly used metric Absrel (Absolute relative difference)\cite{gurram2023metrics}. The absolute relative difference is given by $\frac{1}{H\times W} \sum_{i\in H\times W} |y_{i}-y_{i}^{*}|/y_{i}^{*}$, where $y$ and $y^{*}$ are the predicted and ground-truth depth, respectively.

\subsection{Baseline Comparisons}       
The results of pushing and pulling are shown in Tab. \ref{tab:push} and \ref{tab:pull}, and the long-term experiments are denoted as Ours (long).
The long-term experiment \textit{Ours(long)} demonstrates the effectiveness of force-feedback controller in our manipulation system.
Following \cite{Mo_2019_CVPR}, we conduct the comparisons with other approaches only on short-term interactions since this can reflect how well the model can perform given the initial state of the object, which is the preliminary condition in realizing the whole long-term manipulation procedure.
The average Absrel for the refined depth map on all seen categories is 0.2, which is improved compared with initial depth map of 0.5 Absrel. Moreover, the average Absrel for the refined depth map on all unseen categories is 0.35, showing its domain generalization ability.

\begin{table*}[tb]
\caption{Comparisons of our method under the pushing primitive and ablation study.} \vspace{-0.2cm}   
		\label{tab:push}
	\begin{center}

	\small
	    \setlength{\tabcolsep}{1mm}{
		\begin{tabular}{c|c|c|ccccccccc|c|ccccc}
			\hline
	\multirow{2}{*}{\textbf{}}&\multicolumn{11}{c}{\textbf {Novel Instances in Train Categories}}\vline& 
			\multicolumn{6}{c}{\textbf {Test Categories}}\\
			
	Method
 & Obs.
        &\textbf {AVG}
        &\includegraphics[width=0.03\linewidth]{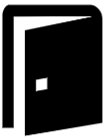}
        &\includegraphics[width=0.03\linewidth]{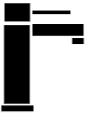}
        &\includegraphics[width=0.03\linewidth]{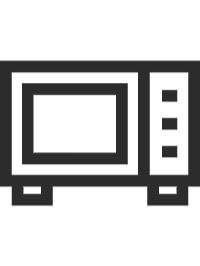}
        &\includegraphics[width=0.03\linewidth]{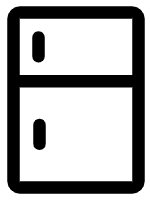}
        &\includegraphics[width=0.03\linewidth]{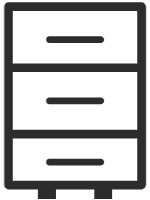}
        &\includegraphics[width=0.03\linewidth]{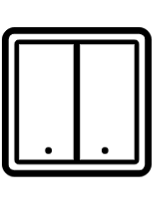}
        &\includegraphics[width=0.03\linewidth]{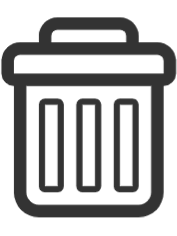}
        &\includegraphics[width=0.03\linewidth]{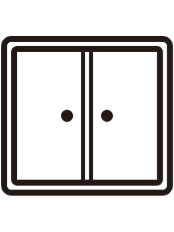}
        &\includegraphics[width=0.03\linewidth]{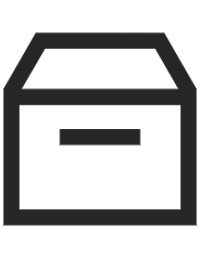}
        &\textbf {AVG}
        &\includegraphics[width=0.03\linewidth]{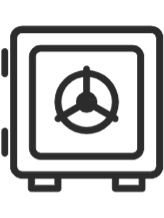}
        &\includegraphics[width=0.03\linewidth]{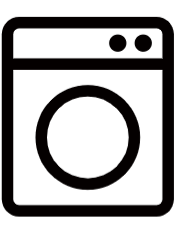}
        &\includegraphics[width=0.03\linewidth]{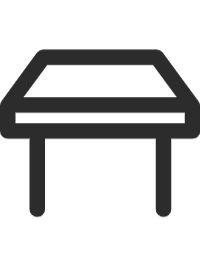}
        &\includegraphics[width=0.03\linewidth]{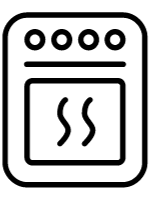}
        &\includegraphics[width=0.03\linewidth]{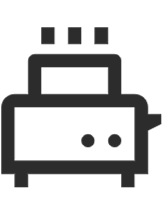}\\\hline\hline
  Where2Act\cite{Mo_2019_CVPR}& PC & 0.36&	0.27&0.46&0.56&0.38&0.26&0.17&	0.42&0.21&0.55&0.21&0.25&0.26&0.18&0.21&0.14 \\
  
  UMPNet\cite{xu2022universal}&PC & 0.69&0.64&\textbf{0.90}&0.65&0.70&\textbf{0.80}&\textbf{0.82}&\textbf{0.84}&0.15&0.60&0.39&0.25&0.33&\textbf{0.77}&0.28&0.32\\
  FlowBot3D\cite{eisner2022flowbot3d}& PC & 0.63&0.54&0.80&\textbf{1.0}&\textbf{0.96}&\textbf{0.80}&0.14&0.69&0.29&0.40&0.29&0.42&0.3&0.18&0.35&0.20 \\ \hline
  
 DrQ-v2\cite{yarats2021mastering}& RGB & 0.20&0.26&0.07&0.27&0.29&0.11&0.11&0.12&0.42&0.15&0.11&0.12&0.08&0.10&0.15&0.10 \\
  Look Closer\cite{jangir2022look}& RGB & 0.19&0.28&0.13&0.30&0.18&0.15&0.16&0.10&0.16&0.26&0.10&0.07&0.11&0.12&0.06&0.12 \\ 
  \rowcolor{gray!20}
  Ours& RGB & \textbf{0.73}&\textbf{0.72}&	0.87&0.83&0.83&0.62&0.73&0.69&\textbf{0.68}&\textbf{0.66}&\textbf{0.49}&\textbf{0.46}&\textbf{0.45}&0.58&\textbf{0.66}&\textbf{0.30}\\ \hline 
  \rowcolor{gray!20} Ours (long)& RGB & 0.70&0.71&0.87&0.83&0.84&0.68&0.59&0.76&0.46&0.55&0.47&0.47&0.41&0.55&0.62&0.29\\\hline\hline
   w/ pixel fusion& RGB & 0.66&0.68&0.65&0.65&0.79&0.72&0.68&	\textbf{0.74}&0.52&0.50&0.32&0.34&0.36&0.40&0.28&0.22 \\ 
   w/o next view& RGB & 0.32&0.27&0.56&0.53&0.22&0.23&0.32&0.36&0.14&0.25 &0.20&0.15&0.23&0.26&0.19&0.15 \\ 
   w/ random next view& RGB & 0.70 & \textbf{0.75}&0.77&0.73&0.82&\textbf{0.72}&0.60&0.73&\textbf{0.70}&0.50 & 0.44&0.42&0.42&0.58&0.63&0.29\\ 
	   w/ depth GT& RGB & 0.77&0.77&0.89&0.85&0.87&0.84&0.55&0.69&0.85&0.60 & 0.51&0.43&0.5&0.57&0.68&0.38 \\ 
     w/ more next views& RGB & 0.76&0.76&0.85&0.87&0.87	&0.61&0.75&0.74&0.72&0.65 & 0.50&0.47&	0.44&0.59&0.66&0.36 \\ \hline 
		
		\end{tabular}}
		
	\end{center}
\end{table*}
\begin{table*}[tb]
\caption{Comparisons of our method under the pulling primitive.} \vspace{-0.2cm}   
		\label{tab:pull}
	\begin{center}
	
	\small
	    \setlength{\tabcolsep}{1.5mm}{
		\begin{tabular}{c| c | c |c c c c c|c| c c c c c}
			\hline
	\multirow{2}{*}{\textbf{}}&\multicolumn{7}{c}{\textbf {Novel Instances in Train Categories}}\vline&
			\multicolumn{6}{c}{\textbf {Test Categories}}\\
			
	Method
    & Obs.
        &{\textbf {AVG}}
        &\includegraphics[width=0.03\linewidth]{./images/mini_img/door.png}
        &\includegraphics[width=0.03\linewidth]{./images/mini_img/faucet.png}
        &\includegraphics[width=0.03\linewidth]{./images/mini_img/fridge.png}
        &\includegraphics[width=0.03\linewidth]{./images/mini_img/storage.png}
        &\includegraphics[width=0.03\linewidth]{./images/mini_img/microwave.png}
        &{\textbf {AVG}}
        &\includegraphics[width=0.03\linewidth]{./images/mini_img/table.png}
        &\includegraphics[width=0.03\linewidth]{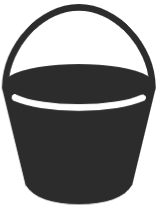}
        &\includegraphics[width=0.03\linewidth]{./images/mini_img/safe.png}
        &\includegraphics[width=0.03\linewidth]{./images/mini_img/washing.png}
        &\includegraphics[width=0.03\linewidth]{./images/mini_img/toaster.png}\\\hline\hline
  Where2Act\cite{Mo_2019_CVPR}& PC & 0.23&0.26&0.33&0.21&0.15&0.18&0.12&0.08&0.16&0.12&0.15&0.10 \\ 
  UMPNet\cite{xu2022universal}&  PC & 0.22&0.18&0.17&0.16&0.31&0.26&0.15&0.15&0.13	&0.16&0.23&0.08 \\
  FlowBot3D\cite{eisner2022flowbot3d}&  PC & 0.30&0.13&0.37&0.47&0.18&0.33 &0.17&0.18&0.15&0.14&0.26&0.12\\\hline
    DrQ-v2\cite{yarats2021mastering}& RGB & 0.07&0.03&0.06&0.05&0.11&0.08&0.06&0.03&0.04&0.06&0.08&0.11\\
  Look Closer\cite{jangir2022look} & RGB & 0.05&0.04&0.03&0.07&0.05&0.04 & 0.05&0.03&0.05&0.08&0.05&0.06\\
  \rowcolor{gray!20}
  Ours& RGB & \textbf{0.51} & \textbf{0.47}&\textbf{0.55}&\textbf{0.36}&\textbf{0.66} & \textbf{0.51} & \textbf{0.26}& \textbf{0.25}	&\textbf{0.32}&\textbf{0.16}&\textbf{0.27}&\textbf{0.30}\\ \hline 
  \rowcolor{gray!20} Ours (long)& RGB & 0.48&	0.47&0.55&0.33&0.56&0.47&0.22&0.23&0.20&0.14&0.24&0.29\\\hline
		\end{tabular}}
		
	\end{center}
\end{table*}

For a fair comparison, we compare with base methods under the same training dataset and end-effector. 
Though point-cloud based methods show promising performance, they rely on the cost-ineffective depth perceiver. Meanwhile, most failure cases of these approaches come from when the robot attempts to interact with the very edge of the object but fails to contact the object because the edge is blurred in the point-cloud modality. 
For RGB-based methods, they all adopt RL-based learning strategy. We notice that these methods had a huge performance drop on unseen shapes within the seen categories as well as unseen categories, potentially due to the weakness in the generalizability of pure RL methods.

\subsection{Ablation and Analysis}
\label{sec:abla}
We provide ablation studies in the bottom part of Tab. \ref{tab:push} to verify the effectiveness of each proposed module.

\textit{With pixel fusion:}
To demonstrate the effectiveness of the token-wise fusion module, we provide the straightforward pixel-level fusion alternative that fuses the high-dimensional feature of the view with their pixel correspondence. Compared to the fusion utilizing pixel-wise dense correspondences (with an average manipulation accuracy of 0.66), token-wise correspondence yields a substantial performance improvement of 0.73 due to its capacity to tolerate depth inaccuracies. 
Furthermore, when contrasting our method against using depth ground truth to establish token-wise correspondence (achieving 0.74), we observe only marginal improvement.
This shows the ability of our fusion module to accommodate depth inaccuracies. Meanwhile, we observe that feature fusion will not directly impact the upper bound of manipulation accuracy.

\textit{Without next view:}
We verify the effectiveness of introducing the next view by conducting training and testing only with the global visual prior, \emph{i.e.}, initial depth map, and initial affordance map. We observe that performance drops dramatically because the global view may only capture a partial observation. The missed crucial details and incomplete object representation can not ensure a reasonable robotic manipulation prediction. We further show the visualization comparisons of initial and refined affordance map on the project webpage.

\textit{With random next view:}
To verify the effectiveness of our best view selection module, we introduce a random next view selection mechanism. Given the inital affordance and depth map, it randomly selects a candidate to place camera among the nine candidates. In this case, the manipulation success rate drops slightly since the randomly collected view may not provide valuable information to complement the global view. Meanwhile, more observations still help in formulating a comprehensive object representation for a thorough manipulation decision. More visualization of the refined map given all candidates and the upper bound analysis of view selection module is presented on the website.

\textit{With more next views:}
In the main experiments, we only adopt two views, one global view and one best next view. In this part, we conduct experiments on three views, with one global view and two consecutively selected next best views, to explore the effectiveness of the number of views. As shown in Fig. \ref{tab:push}, increasing viewpoints can boost the performance by a small margin since more views bring complementary details, facilitating a reliable manipulation inference. However, introducing more views also brings additional computational costs. As a trade-off between accuracy and efficiency, we assume the two views are an optimal option. More analysis regarding space scope is presented on the webpage.

\textit{With depth ground-truth:}
Ours with depth ground-truth (GT) replaces the depth prediction with depth GT. By doing so, we aim to reflect on whether the inaccuracy of depth prediction will directly impact robotic manipulation. We find that with depth ground-truth, the manipulation success rate is only improved by a small amount because
depth ground-truth can help to ensure a relatively accurate contact with the object. 
To mitigate the effects of inaccurate depth estimation, we adopt a self-correction mechanism.
Following the manipulation policy of Flowbot3D \cite{eisner2022flowbot3d}, which detects contact between the gripper and the object, we also assess whether there is contact between the gripper and the object based on the depth prediction. 
If no contact is detected, we gradually adjust the gripper's position along the predicted orientation until contact is established. 
This approach reduces the impact of inaccurate depth estimation on the manipulation process. Consequently, we can conclude that as long as the accuracy of depth estimation falls within a reasonable range, it will not heavily impact the final manipulation outcomes.
Furthermore, by predicting depth, our model implicitly extracts the geometric structure of objects, making informed decisions regarding 3D object manipulation. In contrast, the alternative approach of directly using depth ground truth only captures 2D features without a comprehensive understanding of object geometry. This explains why, in some categories, depth prediction can even outperform depth ground truth in terms of manipulation performance.

\subsection{Real-world Experiment}
\begin{figure}[h]

    \begin{center}
        \includegraphics[width=\linewidth]{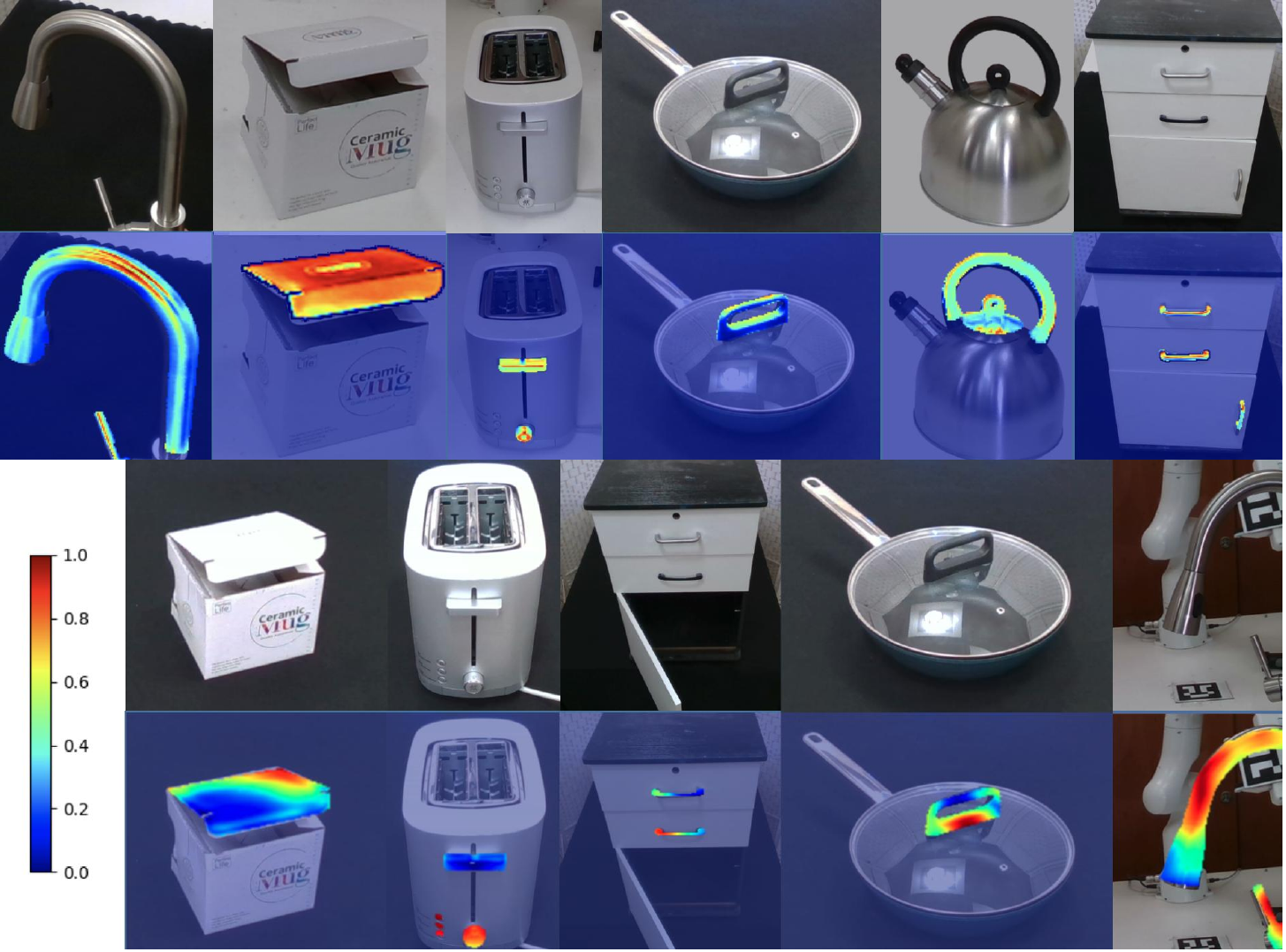}
    \end{center}
    \caption{The upper pairs represent refined affordance map, representing the probability of successful manipulation at that point. The bottom pairs stand for refined depth map, representing normalized depth distances.}
    \label{fig:real-world-exp}
\end{figure}

We conduct experiments that involve interacting with various real-world household objects. 
We employ a Franka Emika robotic arm with a RealSense 415 camera (the only camera we have), but only use RGB images as inputs. 
To achieve sim-to-real transfer, we've devised two strategies:
1) Pretraining Strategy: For the visual encoder, we've incorporated ImageNet pre-trained weights into the ResNet architecture. This integration equips the model with strong real-world perception capabilities. 
2) Simulator Strategy: In simulator, we've employed domain randomization to amplify scenario diversity, diversifying elements like lighting, materials, light position, etc,  aiming to ease sim-to-real transfer. We visualize the domain randomization of object material along with the manipulation process video on the project website.
Since manipulation performance relies predominantly on the quality of the affordance and depth prediction, we demonstrate these under real-world settings in Fig. \ref{fig:real-world-exp}
, showing the effectiveness of our model in real-world.

\vspace{-0.05cm}
\section{Conclusion and Limitation}
In this paper, we investigate image-based robotic manipulation learning with the acquisition of depth and affordance. A global view is utilized to provide a global visual prior for subsequent viewpoint selection. Furthermore, we introduced the best next view selection and geometric-consistency fusion to extract a refined visual prior for accurate manipulation prediction. 
We believe generalizable image-based robotic manipulation is an exciting direction because RGB data can be captured at a low cost which will be beneficial to low-cost robot systems in the future. 
As for the limitation, we limit out experiments to two types of action primitives with individual initial motion trajectories. In the future, we plan to generalize the framework to more free-form interactions.
{
\bibliographystyle{IEEEtran}
\bibliography{IEEEabrv,reference}
}

\end{document}